\documentclass[submission]{eptcs}
\providecommand{\event}{PMEA} 
\usepackage{breakurl}             
\usepackage[pdftex]{graphicx}
\usepackage{titlesec}
\usepackage{amsmath}
\usepackage{float}
\usepackage[inline]{enumitem}
\usepackage[tight]{subfigure}
\usepackage[algo2e,ruled,vlined]{algorithm2e}
\usepackage{harvard}

\expandafter\def\expandafter\normalsize\expandafter{%
	\normalsize
	\setlength\abovedisplayskip{10pt}
	\setlength\belowdisplayskip{10pt}
	\setlength\abovedisplayshortskip{20pt}
	\setlength\belowdisplayshortskip{20pt}
}

\title{Probabilistic Model-Based Approach for Heart Beat Detection}

\author{Hugh Chen
\institute{University of California Berkeley\\
California, USA}
\email{hugh.chen\{at\}berkeley.edu}
\and
Yusuf Erol
\institute{University of California Berkeley\\
California, USA}
\email{yusufbugraerol\{at\}berkeley.edu}
\and
Eric Shen
\institute{University of California Berkeley\\
	California, USA}
\email{ericshen\{at\}berkeley.edu}
\and
Stuart Russell
\institute{University of California Berkeley\\
    California, USA}
\email{russell\{at\}cs.berkeley.edu}
}
\def\titlerunning{Probabilistic Model-Based Approach for Heart Beat Detection}
\def\authorrunning{Hugh Chen, Yusuf Erol, Eric Shen \& Stuart Russell}
\begin{document}
\maketitle

\begin{abstract}
Nowadays, hospitals are ubiquitous and integral to modern society.  Patients flow in and out of a veritable whirlwind of paperwork, consultations, and potential inpatient admissions, through an abstracted system that is not without flaws.  One of the biggest flaws in the medical system is perhaps an unexpected one: the patient alarm system.  One longitudinal study reported an $88.8\%$ rate of false alarms, with other studies reporting numbers of similar magnitudes.  These false alarm rates lead to a number of deleterious effects that manifest in a significantly lower standard of care across clinics.

This paper discusses a model-based probabilistic inference approach to identifying variables at a detection level.  We design a generative model that complies with an overview of human physiology and perform approximate Bayesian inference.  One primary goal of this paper is to justify a Bayesian modeling approach to increasing robustness in a physiological domain.

We use three data sets provided by Physionet, a research resource for complex physiological signals, in the form of the Physionet 2014 Challenge set-p1 and set-p2, as well as the MGH/MF Waveform Database.  On the extended data set our algorithm is on par with the other top six submissions to the Physionet 2014 challenge.
\end{abstract}

\smallskip
\noindent \textbf{Keywords:} Beat Detection, PhysioNet Challenge, Particle Filter, Dynamic Bayesian Network, ECG, Blood Pressure, Model Based Probabilistic Inference.
\newpage

\section{Introduction}
Patient monitoring is a significant part of health care, not only to ensure that physicians can accurately diagnose and treat patients, but also to trigger biometric-based alarms.  The intent of these alarms is to guarantee that a patient receives attention from clinicians whenever his or her condition takes a turn for the worse.

One of the important facets of such biometric data are heart beats.  By monitoring heart beats, a number of cardiac issues, including a variety of life-threatening arrhythmia (asystole, bradycardia, tachycardia, etc.), can be detected \cite{PhysioNet}.  In the case where there is a high noise level and poor signal quality the heart beats themselves are unreliable.  These unreliable beats can lead to false-negatives, where alarms fail to be triggered, as well as false-positives, where alarms are triggered for no substantive reason.  These two cases respectively result in alarm failure and alarm fatigue \cite{alarmarticle}.  

False-negatives are direct alarm failures.  Instances of alarm failure are dangerous because they mean that patients in life-threatening situations may be completely overlooked.  Alarm fatigue, on the other hand, is an indirect consequence of an excess of false-positives (AKA false alarms).  An excess of false alarms has a number of pernicious effects, including the desensitization of nurses and doctors to true alarms, disturbance of ailing patients, and the cost of time wasted for both physicians and patients.  One article cites alarm fatigue as one of the top patient safety concerns in hospitals \cite{AlarmSignificance}.  In a 31-day study across 461 adults in intensive care units, 88.8\% of the 12,671 arrhythmia alarms were falsely detected \cite{AlarmFatigueArticle}.  Other studies report similar numbers, indicating that alarm fatigue is a real phenomenon.

Typically, the methods used to identify the heart beats are straightforward signal processing algorithms that take advantage of the regular waveforms of either electrocardiogram (ECG) or arterial blood pressure (ABP) signals.  In ECG signals, there exists a regular QRS complex.  In signals with well-defined QRS complexes, executing signal processing algorithms that annotate heart beats at the R peaks achieves high accuracy.  Likewise for the ABP signals, there exists a spike, albeit not as sharply defined as the QRS peak, that indicates the location of the heartbeat.  In ABP, heart beat detection is further complicated by a delay between heartbeats and the pressure peaks.
\begin{figure}[ht]
	\centerline{
		\includegraphics[width=0.6\textwidth]{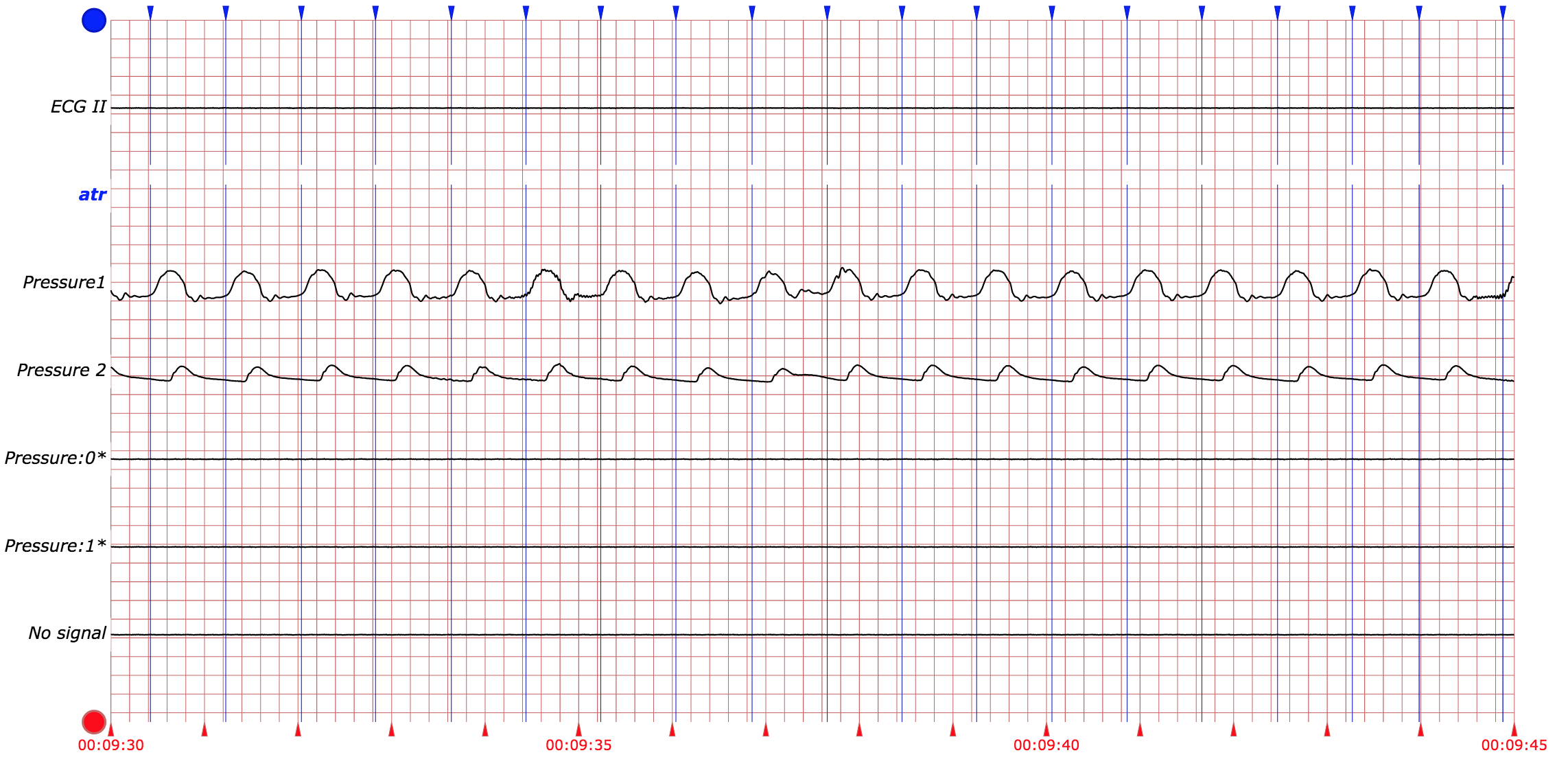}
		\includegraphics[width=0.6\textwidth]{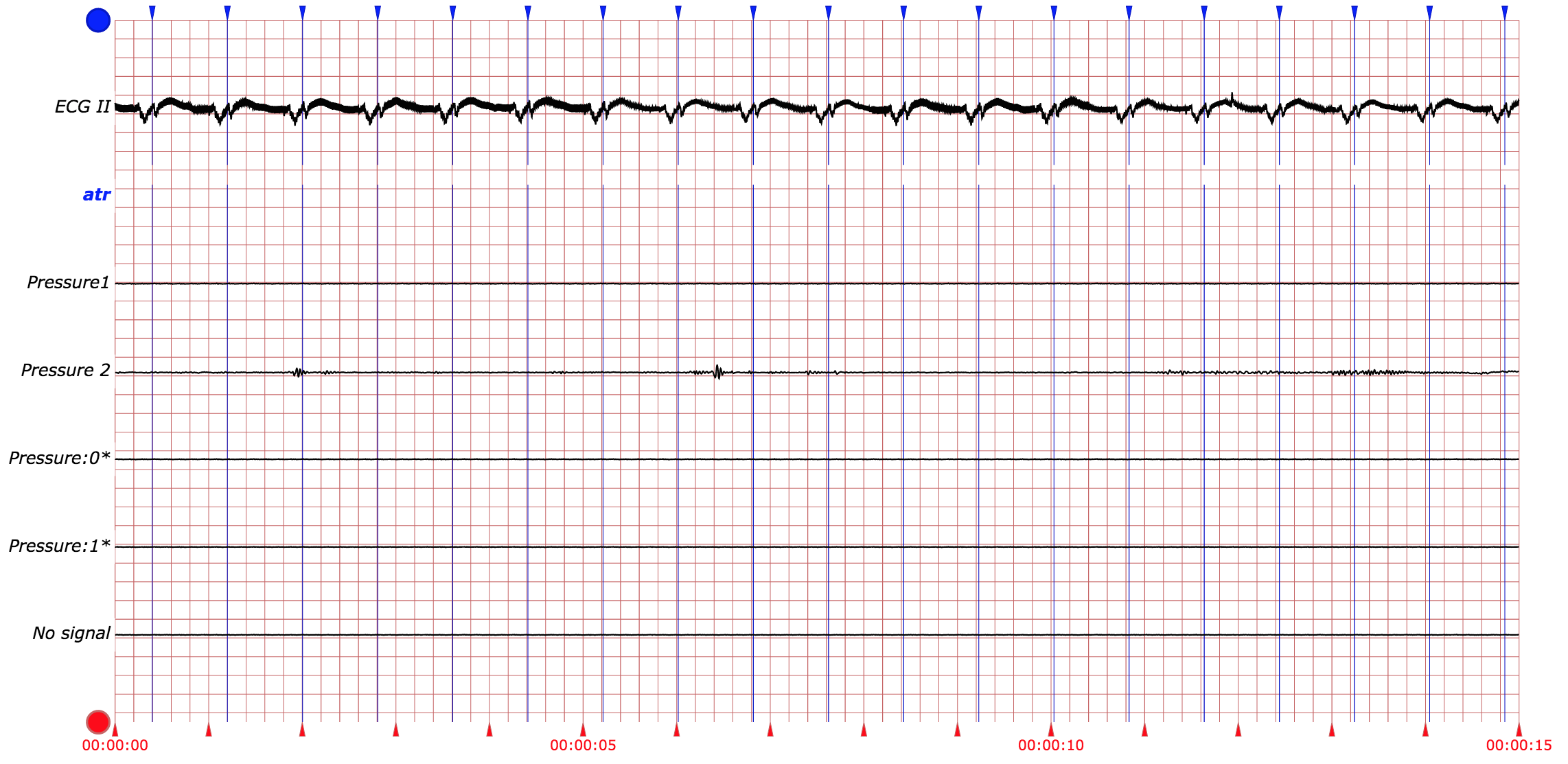}
	}
	\caption{Section of signal from example 1522 of Physionet 2014 Challenge's set-p2.  In the left subsection you can see that the electrocardiogram has flat-lined.  On the right, we conversely see an example where blood pressure has flat-lined, all within the same signal.}
	\label{DroppedSignalExample}
\end{figure}
Standard methods for heart beat detection typically include signal processing algorithms that rely on a particular lead for a particular signal.  This naive approach is already problematic because of the potential for dropped signals.  Figure \ref{DroppedSignalExample} provides an example of dropped signals within a single piece of data.  This shows that for a given patient, across a relatively short period of time, it is possible for either the ECG or the ABP signal to simply flat-line and yield no useful information.  These dropped signals occur for a variety of reasons, including but not limited to technical malfunctions or the detachment of sensors due to patient movement.  In these cases, any signal processing algorithm that naively depends on a single signal's lead will fail to provide useful data for extended periods of time.  

The problem of dropped signals almost naturally suggests a solution in the form of a signal switching algorithm.  One that evaluates whether the lead has flatlined, and if so, switches to a lead with a notable stream of data.  This approach certainly works to a degree, but a naive application will fail due to events that disturb the signal and generate noise, otherwise known as artifacts.
\begin{figure}[ht]
	\centerline{
		\includegraphics[scale=.3]{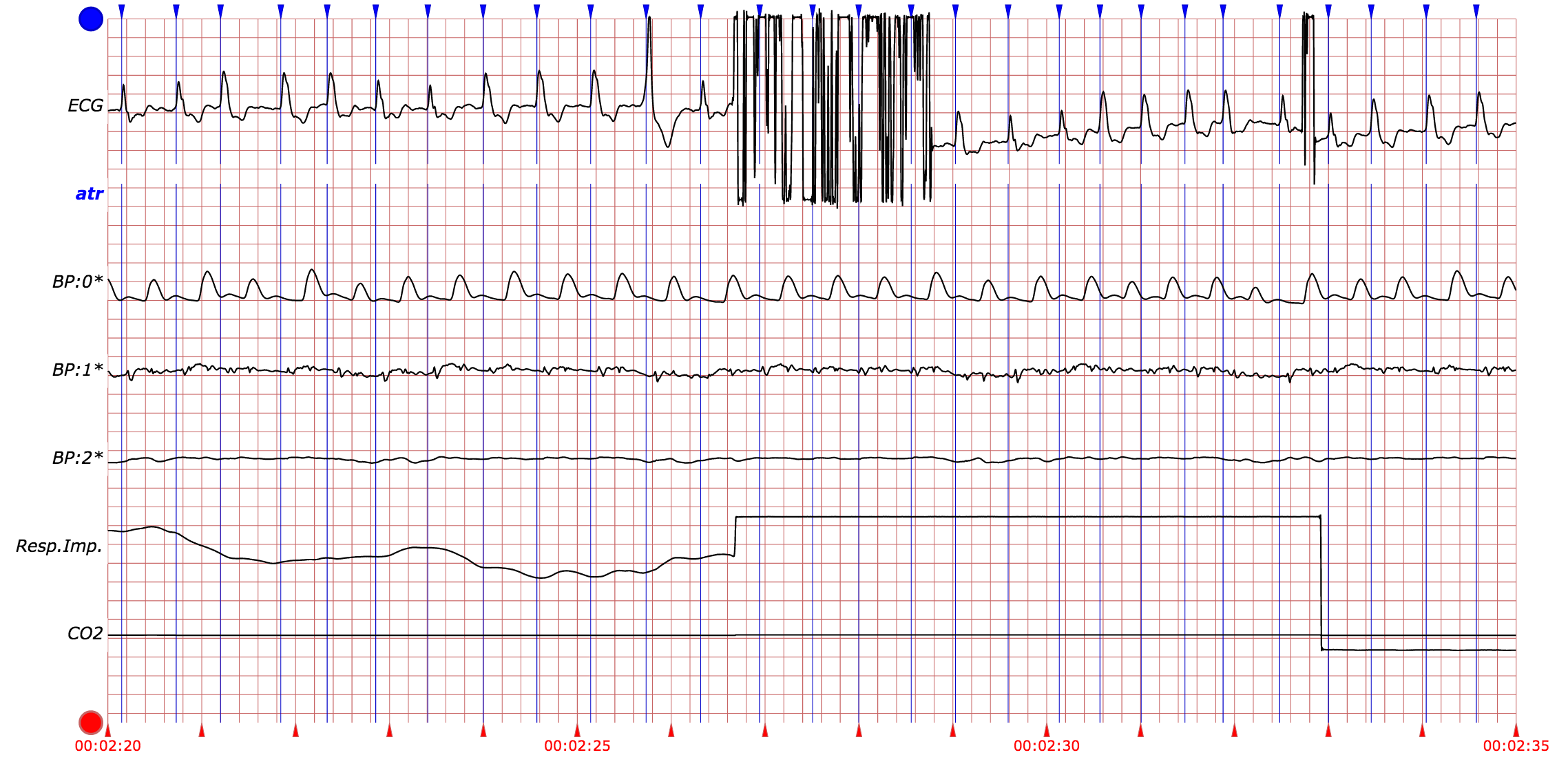}
	}
	\caption{Section of signal from example 1715 from Physionet 2014 Challenge's set-p2.  In this subsection you can see that the electrocardiogram has an area with a significantly jittery signal.}
	\label{ArtifactExample}
\end{figure}
In Figure \ref{ArtifactExample}, the middle of the ECG signal has a clear example of an artifact.  The regular QRS complexes are disturbed, leaving a signal that bears no recognizable patterns.  Artifacts can be the manifestation of a patient brushing their teeth, bumping into something, or simply rolling around in their sleep.  They complicate the detection of heart beats, because the artifacts can corrupt the signals in a variety of ways, thereby rendering any naive switching algorithm insufficient.  Ultimately, the goal is to robustly detect heart beats despite the occurrence of artifacts, noise, and dropped signals.  In doing so, patient monitoring systems in hospitals can become much more efficient, reducing ICU false alarm rates.

Given all of this clinical data, we recognize the problem as one that can be solved through either a data-driven or a model-based approach.  We could potentially treat the problem as a regression problem and allow the program to train on the Physionet data and develop its own interpretation and understanding of the problem.  Alternatively, we can assume that the problem is an inherently biological problem, and take an approach that borrows from modern understanding of human physiology.  

Because there exists a large corpus of research in the direction of human physiology, we choose the latter.  Yet earlier, we have recognized that the data we are dealing with is inherently uncertain.  In order to capture both the model and inherent uncertainty we approach the problem through a model-based probabilistic inference approach.  

We develop a Dynamic Bayesian Network (DBN) to describe the interactions of the patient's physiological characteristics over time.  In order to incorporate our beliefs about how artifiacts and noise manifest themselves in data we also incorporate an observation model.  Finally, we use particle filtering, which is a Sequential Monte Carlo (SMC) method, to perform combined state and parameter estimation on our non-linear, non-Gaussian model.

\section{Physionet}
Before moving directly into the algorithm it is worth mentioning Physionet.  Physionet is a research resource that offers access to a sizable supply of recorded physiological signals and their open source software.  In addition, they hold challenges on a yearly basis, with last year's challenge pertaining to heart beat detection.  We use several of the datasets provided by Physionet to evaluate our algorithm's performance.  

\subsection{Material}
The first is the Physionet Challenge 2014 training set (set-p1), which is was a dataset released during the first phase of the challenge \cite{PhysioNet}.  There are 100 examples that are all relatively clean and artifact-free.  The data is typically 10 minutes long or shorter and each record contains at least one ECG signal and at least one ABP signal.  The sampling frequency is also consistent among these examples at 250 samples per second.  

The next dataset set is the Physionet Challenge 2014 extended training set (set-p2), consisting of 100 records \cite{PhysioNet}.  These records contain signals that have more noise and artifacts than those of set-p1.  The signals in set-p2 are mostly 10 minutes long, although there are occasionally shorter signals.  The sampling frequency varies between 250 samples per second to 360 samples per second.

The final dataset we use is the Massachusetts General Hospital/Marquette Foundation (MGH/MF) Waveform Database provided by Physionet \cite{MGHDB}.  The database consists of 250 recordings and represents a broad spectrum of physiologic and pathophysiologic states.  Individual recordings vary in length from 12 to 86 minutes, and in most cases are about an hour long.  The effective sampling frequency is 360 samples per second.

For these three datasets, reference beat annotations are available.  These reference beat annotations represent the consensus of several expert beat annotators, and are used for determining the accuracy of the algorithms.

Beyond the datasets, we also make use of the GQRS and WABP functions, which are the basic beat detectors for ECG and ABP respectively provided by Physionet's WFDB Toolbox \cite{WFDBToolbox}.  In addition, we use other WFDB Toolbox functions for processing signals and annotations.

\subsection{Related Work}
Since Physionet held a challenge and collected many submissions, there are also quite a few bodies of work that make meaningful progress towards improving heart beat detection.  The top six algorithms that were submitted to the Physionet 2014 Challenge, ordered by their performance in Physionet are Pangerc \cite{PangercPaper}, Johnson \cite{JohnsonPaper}, Antink \cite{AntinkPaper}, De Cooman \cite{CoomanPaper}, Johannesen \cite{JohannesenPaper}, and Vollmer \cite{VollmerPaper}.

In general, most of the submissions involved a subset of a few typical techniques: pre-processing, a combination algorithm, and then post-processing.  In addition most algorithms used some form of delay incorporation, either in their pre-processing or combination algorithm to account for the ABP delay.  Finally, a few algorithms make use of signals outside of the ECG and ABP signals.  It is worth noting that Pangerc, whose performance is significantly higher than the other applications involved re-implementing ECG and pulsatile-signal detection algorithms.  For a recent review of the submissions, refer to the Physionet 2014 Challenge summary paper \cite{PMEA}.

\section{Methods - Algorithm}
\subsection{Introduction}
\begin{figure}[ht]
	\centerline{
		\includegraphics[scale=.7]{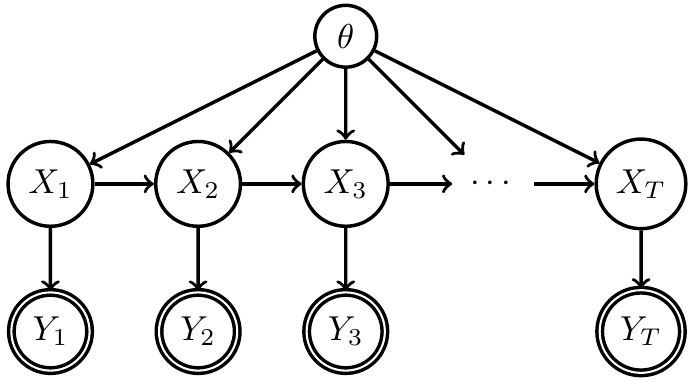}
	}
	\caption{A state-space model with static parameters $\theta$.
$X_{1:T}$ are latent states and $Y_{1:T}$ are observations.}
	\label{StateSpaceModel}
\end{figure}

Dynamic Bayesian networks (DBNs) are widely used to model the processes underlying sequential data such as speech signals, financial time series, genetic sequences, and in our case, physiological signals.  DBNs model a process using static parameters, hidden variables that evolve over time, and observations at each time step, as shown in Figure \ref{StateSpaceModel}.

More specifically, for a partially observable Markov process with unobserved state variables $\left\{X_{t} \right\}_{t \geq 0}$, and observations $\left \{Y_t \right \}_{t \geq 0}$ that is parametrized by a static parameter space $\Theta$, the probabilistic model is defined as follows.
\begin{align}
	X_0 & \sim p(x_0\mid \theta) \\
    X_t \mid x_{t-1} &\sim p(x_t \mid x_{t-1}, \theta) \\
	Y_t \mid x_t &\sim p(y_t \mid x_t, \theta)
	\label{eq:state-space-model}
\end{align} 
The first equation represents the initialization of state variables, which corresponds to the prior probability. The second equation represents the propagation model, which is based on transition probabilities from one state to another. For our propagation model, we implemented a drastically simplified model of human biometrics and then evolved variables in a physiological manner.  The final equation represents the observation model, which corresponds to the probability of a particular observation given a certain state. In our case, our observations consisted of annotations and signal quality indices from signal detection algorithms.

\subsection{Sequential Monte Carlo}

Given a probabilistic model and a sequence of observations, one can attempt to estimate the latent states. That is the problem of state estimation, also known as filtering: the process of computing the posterior distribution of the hidden state variables given a sequence of observations.  Exact filtering is often intractable except for specific cases, but approximate filtering using the particle filter (a sequential Monte Carlo method) is feasible in many applications \cite{Particlefilter} \cite{ParticleLearningAndSmoothing}.

The specific method we use is Sequential Importance Sampling-Resampling (SIR), otherwise known as bootstrap filtering and particle filtering.  This representation relies on approximating the posterior density, $p(x_t|y_{0:t},\theta)$ function at any given time using a set of random particles that we recursively evolve.  As the cardinality of the set grows, the approximation improves in accuracy.

We initialize the states of our particles based on the prior probabilities. We then propagate the state of the particles using the transition probabilities, weight based on the observation probabilities, and resample at each time step.  Through this propagate-weight-resample scheme, particle filtering generates simulations that explore the likely portions of the latent probability space. 

\begin{algorithm2e}[ht]
\KwIn{$N$: number of particles;\\
$y_0,\dots, y_T$: observation sequence}
\KwOut{$\bar{x}_{1: T}^{1: N}$}
initialize $\left\{x_0^i\right\}$ \;
\For{$t=1,\ldots,T$} {
	\For{$i=1,\dots,N$}{
	        sample $x_{t}^i \sim p(x_t \mid x_{t-1}^i)$\;
	        $w_t^i \leftarrow p(y_t \mid x_t^i)$\;
	}
	resample $\left\{\frac{1}{N},\bar{{x}}_t^i\right\}\leftarrow  \left\{w_t^i,{x}_t^i\right\}$\;
	$\left\{{x}_t^i\right\}\leftarrow \left\{\bar{{x}}_t^i\right\}$\;
}
\caption{Sequential importance sampling-resampling (SIR)}
\label{alg:SIR}
\end{algorithm2e}

\section{Methods - Model}
Our approach relies on the assumption that human physiology follows a pattern that can be modeled in a Bayesian manner.  Specifically, we construct a DBN that corresponds to human physiology, with static variables $\theta$ (e.g. resting heart rate), dynamic state variables $X_i$ (e.g. true heart rate) which define a propagation model, and observations $Y_i$ (e.g. ECG annotations and signal quality) which define an observation model.  The propagation and observation models are described in the following sections and illustrated in Figure \ref{DynamicBayesNet}.  We proceed to use the filtering techniques covered in the previous section to perform state estimation on the DBN we have constructed.

\subsection{Propagation Model; $p(x_t|x_{t-1},\theta)$}

\begin{figure}[ht]
	\centerline{
		\includegraphics[scale=.3]{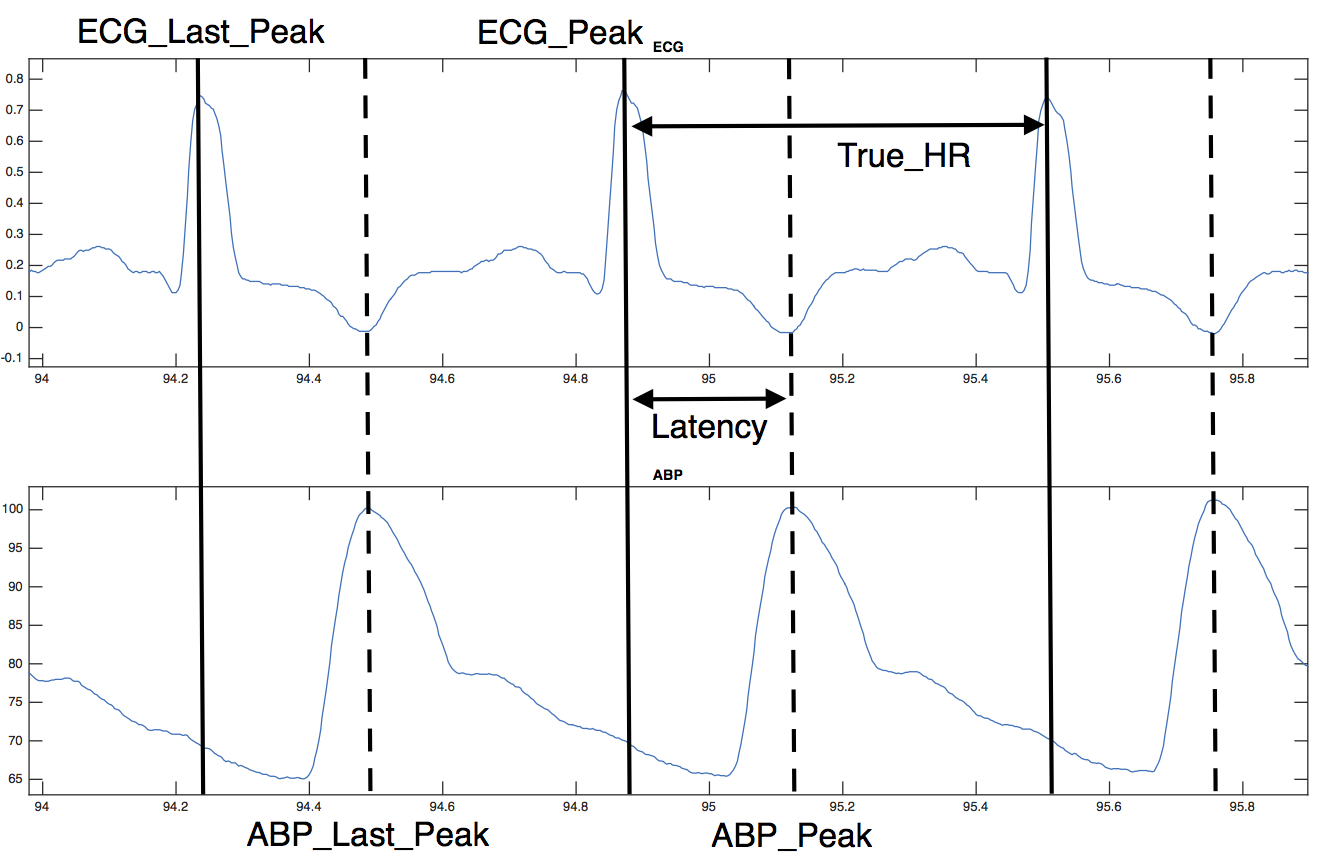}
	}
	\caption{This figure shows the physical representation of some of the propagation variables.}
	\label{PropModelVariables}
\end{figure}

Our propagation model (transition model) encodes a DBN, and indicates our beliefs about the interdependent evolution of our relevant variables over time.  The model makes use of nine variables to represent a simplified model of human physiology.  Refer to Figure \ref{PropModelVariables}, to see the physical representation of a few of the propagation variables.

The first two variables are static parameters: $RestHR$ and $Latency$, which represent the resting heart rate of the patient and the delay between ECG and ABP signals, respectively \cite{Latency}.  These static parameters converge quickly during particle filtering.  

The next variables we cover are latent variables.  The third variable is the $TrueHR$, which represents the belief of the patient's heart rate at a particular point in the signal.  The fourth variable is the $ECGPeak$, which is a binary variable that represents whether there is a peak in the current window of the ECG signal.  The $ECGPeak$ variable evolves based on the $TrueHR$ and the next variable, $ECGLastPeak$, which represents the last time we believed there was a peak in the ECG signal.  The sixth and seventh variables are $ABPPeak$ and $ABPLastPeak$ which are analogous to the corresponding ECG variables, but incorporate the $Latency$ variable.  Note that in this model, the $ECGPeak$ at time $t$ coincides with the $ABPPeak$ at time $t+Latency$.  This means that the $ABPPeak$ variable represents our final belief about heart beat annotations, because it incorporates both ECG and ABP information.  The final two variables are $ECGArtifact$ and $ABPArtifact$, which are binary variables that are used to label a signal as artifactual.  For a more in-depth explanation, refer to the Propagation Model in Appendix \ref{Propagation Model Supplemental}. 

\subsection{Observation Model; $p(y_t|x_t,\theta)$}
The observation model (sensor model) encodes our beliefs about the functions we use to derive observations and the probability that the observations correspond to the current states.  In the observation model there are six variables that we relate to the variables in our propagation model.

The first two variables are annotation observations, $ECGAnn$ and $ABPAnn$.  These binary variables represent whether the algorithms provided by Physionet, GQRS and WABP, found an annotation at the current time.  The observation model describes a relationship between these variables and the corresponding $Peak$, $LastPeak$, and $Artifact$ variables in the propagation model.  The next two are heart rate observations, $ECGHR$ and $ABPHR$, which once again are derived using WABP and GQRS to give an estimate of the heart rate.  These parameters are mainly associated with the $TrueHR$ variable.  Finally we have two SQI observations, $ECGSQI$ and $ABPSQI$, which represent how trustworthy a particular signal is.  The $ECGSQI$ is calculated by comparing the results from two of Physionet's ECG signal detectors, and the $ABPSQI$ is calculated by checking that the detections are within certain physiological boundaries \cite{JohnsonPaper} \cite{SunPaper} \cite{Sunmastersthesis}.  For a more in-depth explanation, refer to the Observation Model in the Appendix \ref{Observation Model Supplemental}.

\begin{figure}[ht]
	\centerline{
		\includegraphics[scale=.3]{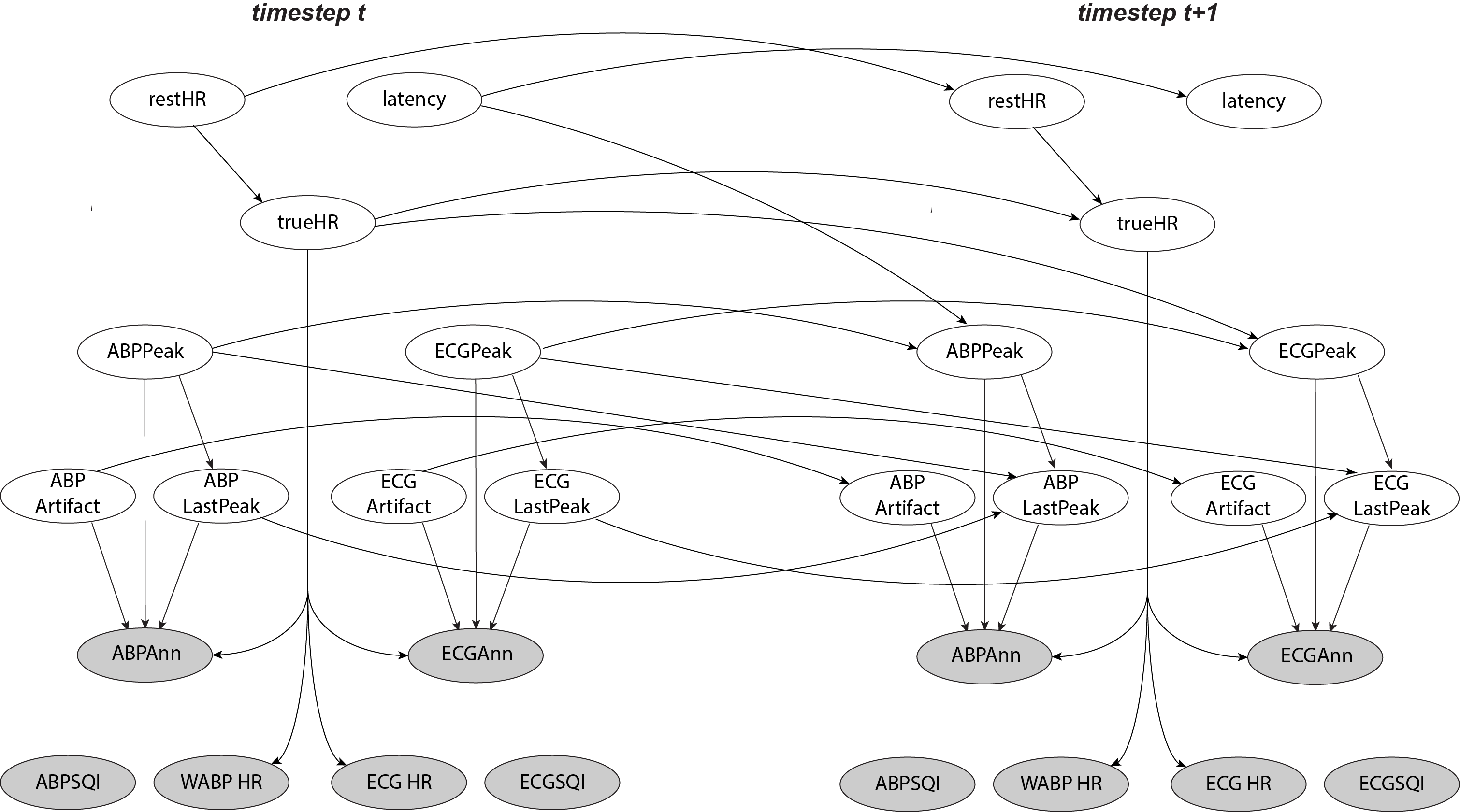}
	}
	\caption{This figure shows the propagation and observation model that encodes our Dynamic Bayesian Network. The grayed out variables correspond to observed variables.}
	\label{DynamicBayesNet}
\end{figure}

\subsection{Performing State Estimation}
Given our probabilistic heart beat model, we now use particle filtering to perform state estimation and to determine when heart beats occurred.  Our particle filter follows SIR whereby we perform three steps recursively: propagation, weighting, and resampling.

First, we split our signals into 25 millisecond windows, and calculate the values of the observation model variables for each of these windows. Then, the particle filter assigns a prior belief to our set of particles (in our algorithm we use a set of 2000).  For the propagation step we propagate the particles individually according to our transition model to acquire the state of the model one step into the future.  Next, for each particle, we calculate a weight which is representative of the likelihood of that particle's realized state given the observations we have made at the corresponding time.  The final recursive step is to resample the particles according to the weights we calculated in order to avoid the \emph{degeneracy problem}, where the particles all have negligible weight \cite{ParticleLearningAndSmoothing}.

We repeat the recursive steps above for the duration of the signal in a sequential fashion.  At each step in the recursion we also save the average state of the variables across the particles, which is used to compose our actual heart beat annotations.  Based on $ABPPeak$s, we backshift a distance of $Latency$ and then annotate a beat only if enough particles are in the state that corresponds to a peak.  This threshold is one of several hyperparameters within our algorithm that were tuned over the development of the filter.  The final annotations we use correspond to timesteps where enough particles were in a state of $ABPPeak$s backshifted by the $Latency$ between $ECGPeak$s and $ABPPeak$s.

\section{Results}
Now that we have established the algorithm, we discuss the performance of the algorithm on several datasets. In this section we compare results in terms of the sensitivity (recall) and positive predictivity (precision).  The sensitivity represents the percentage of actual beats our algorithm annotated, and the positive predictivity represents the percentage of detections that corresponded to actual beats.

\subsection{Results - Individual}

The first example we discuss is record 1376 from set-p2.  This record is an example where we make a large improvement over GQRS and WABP.
\begin{figure}[ht]
	\centerline{
		\includegraphics[scale=.35]{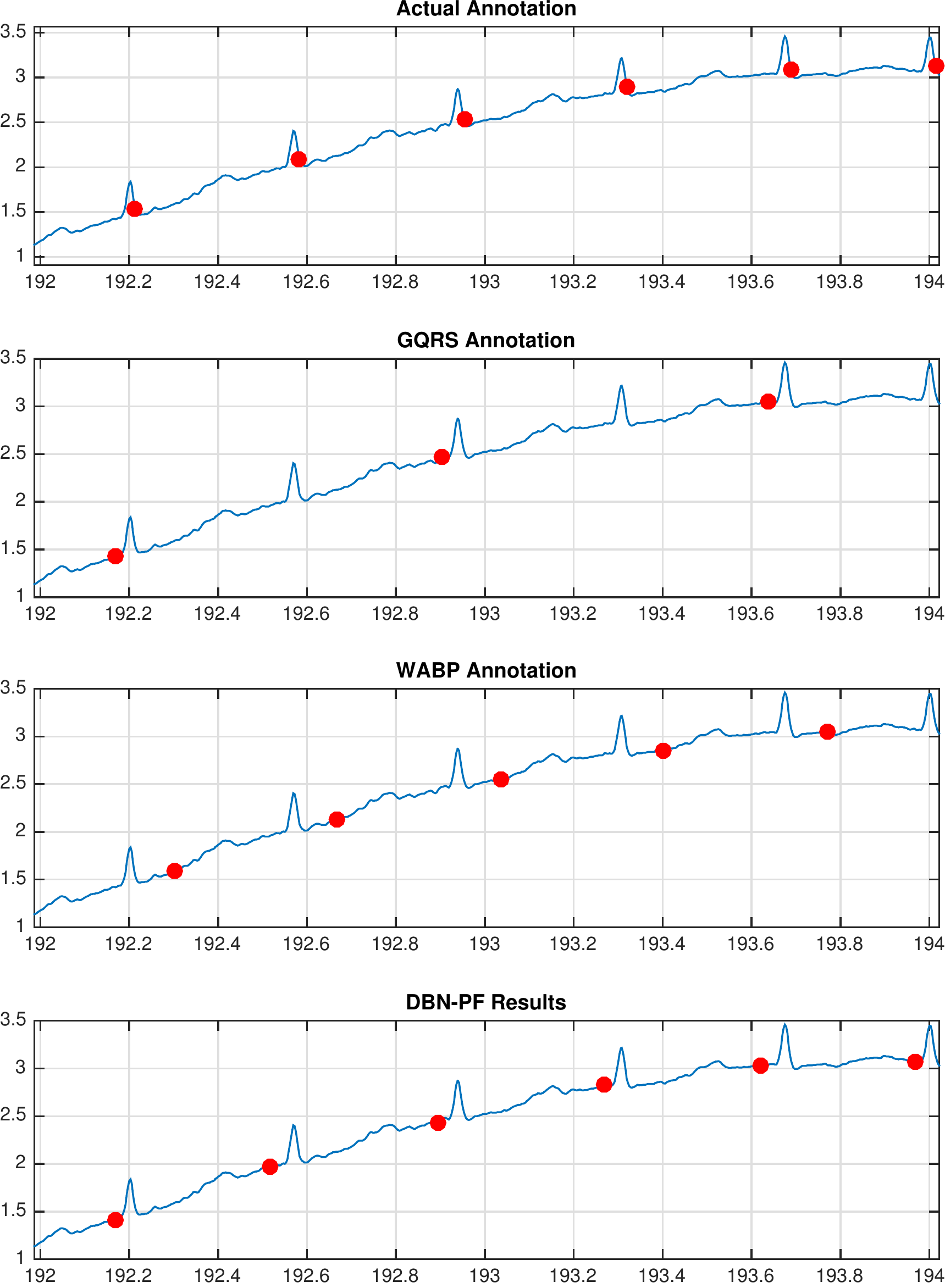}
		\includegraphics[scale=.35]{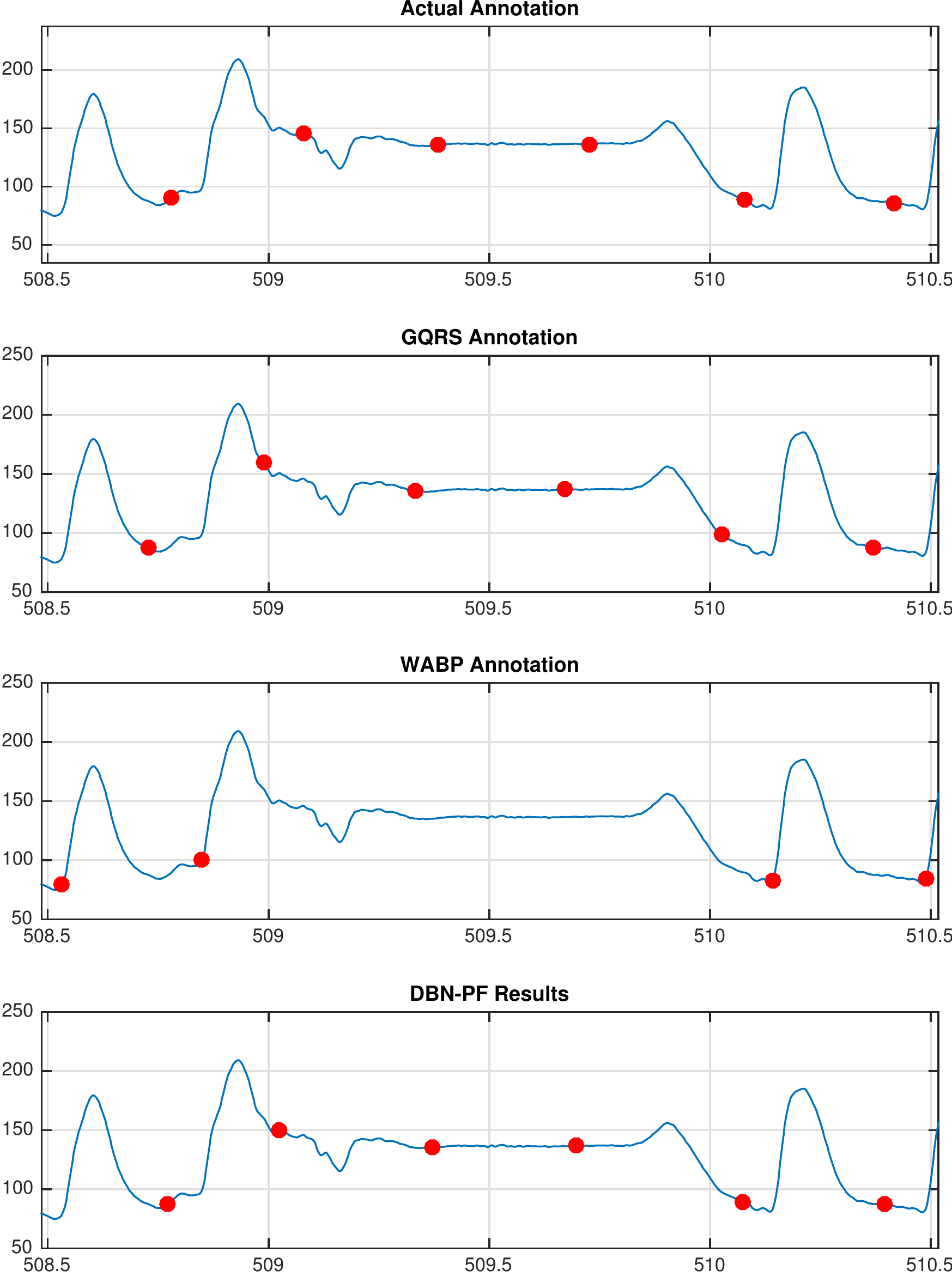}
	}
	\caption{Example 1376 from set-p2.  On the left we plot against the ECG signal, and on the right we plot against ABP signal.  Both signals are plotted over a section of two seconds.  The red dots signify annotations. On this example we achieved a sensitivity of 0.91835 and a predictivity of 0.98842 whereas GQRS had a sensitivity of 0.69051 and a predictivity of 0.99002.  The slightly misaligned beats on the left are likely due to an overestimated latency.}
	\label{1376}
\end{figure}

In Figure \ref{1376} we note that the particle filter can recover from spurious GQRS annotations on the left and spurious WABP annotations on the right, all within the same signal.  Our algorithm performs well overall on record 1376, and this example illustrates its capacity for artifact recovery.  Most of the improvement our algorithm achieves is in knowing when to trust a particular signal to the point that it will incorporate it within our observation model.

The next example we discuss is record 2664 from set-p2.  This record demonstrates our improvement over GQRS and provides an illustrative example of signal quality and artifacts.

\begin{figure}[hp]
\vspace{-0.1in}
\centerline{
\subfigure[ECGSQI]{
	\includegraphics[scale=0.3]{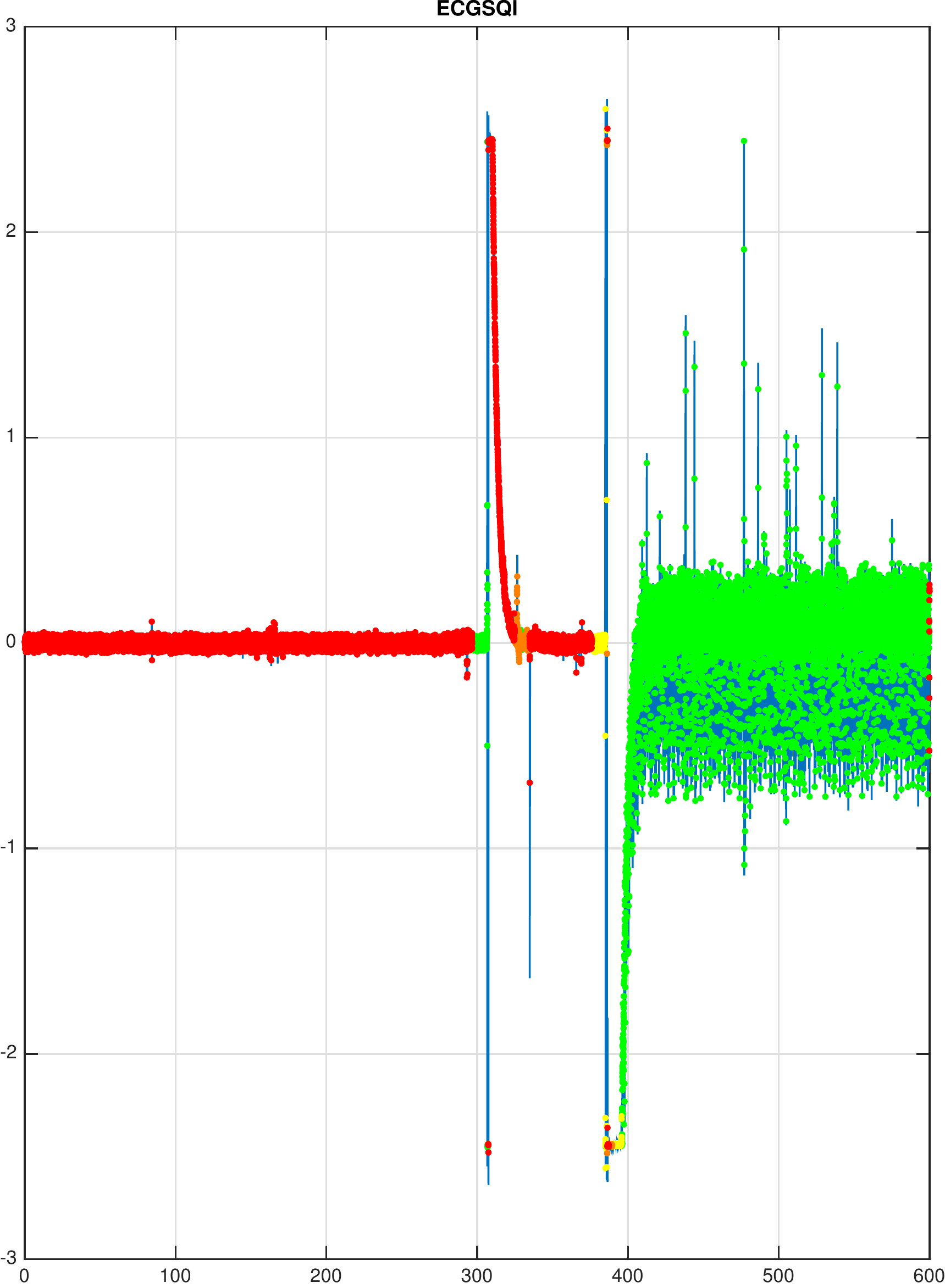}
    \label{ecg_sqi}
}
\subfigure[ECGArt]{
    \includegraphics[scale=0.3]{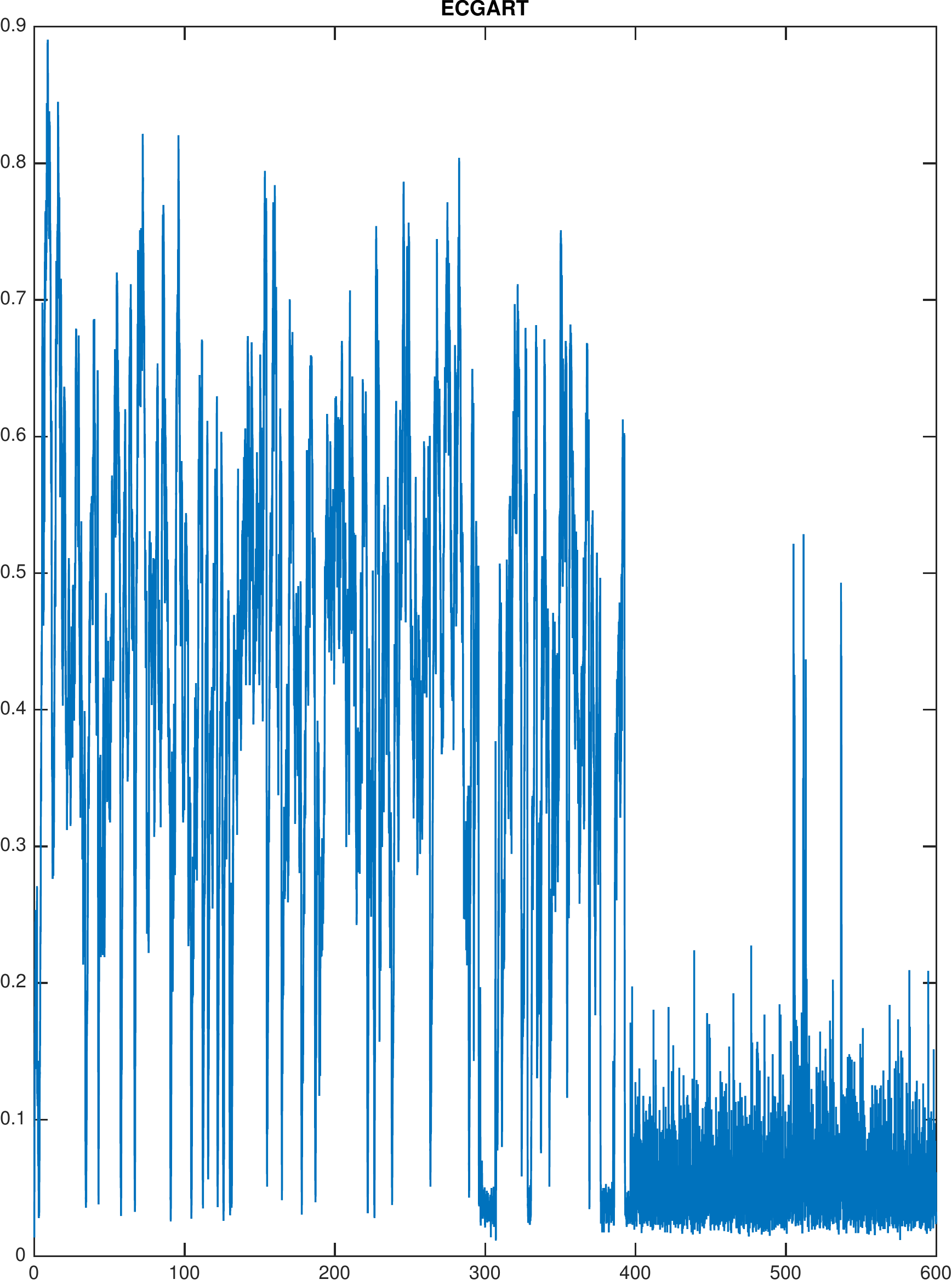}
    \label{ecg_art}
}
\subfigure[ECG HR]{
    \includegraphics[scale=0.3]{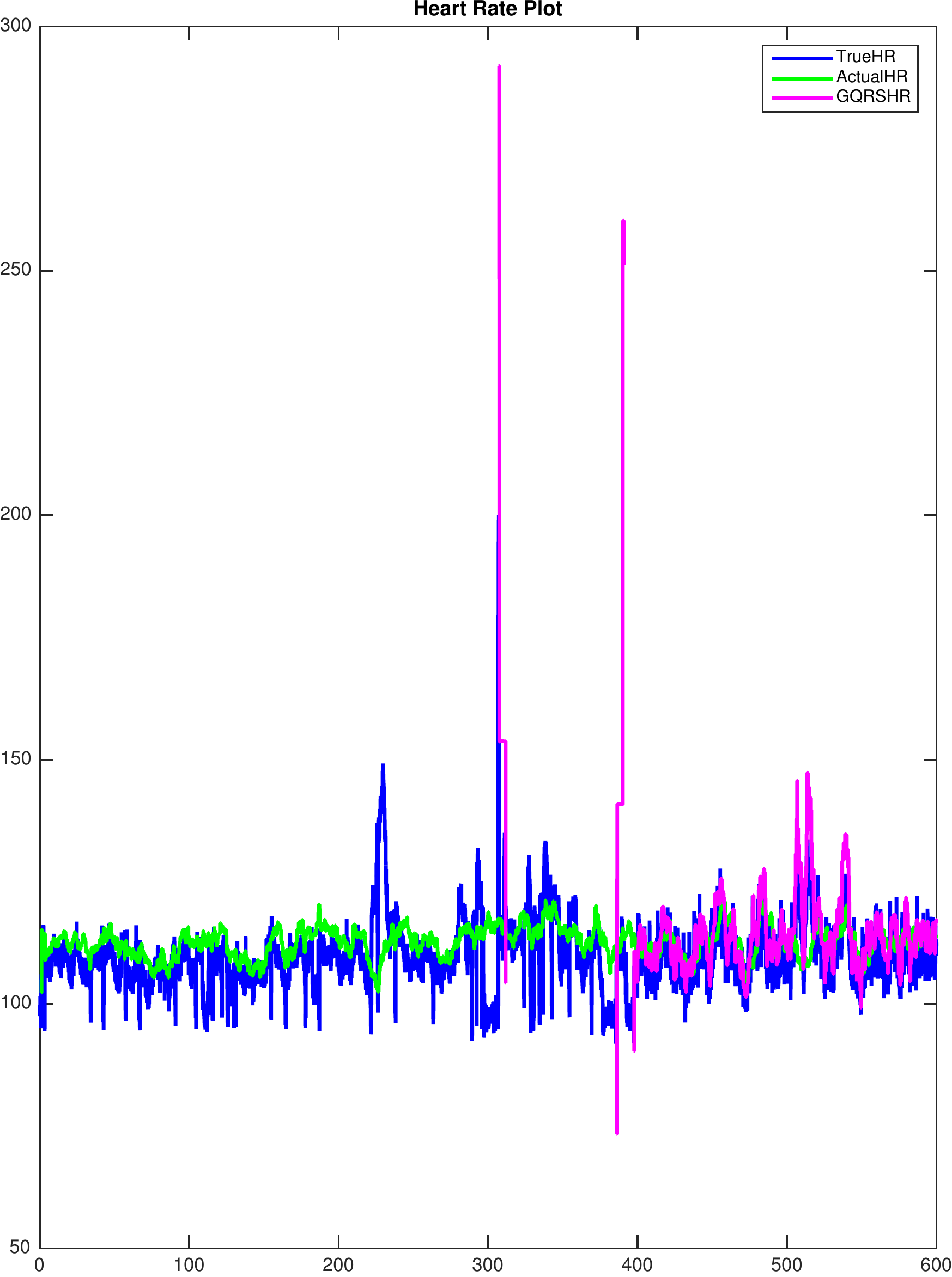}
    \label{ecg_hr}
}
}
\vspace{-0.1in}
\caption{Example 2664 from the set-p2.  On the left we depict the ECG signal quality observation over time.  In the middle we depict the mean of the particles' $ECGArt$ variable over time.  On the right we depict the heart rate over time.}
\label{2664_quality}
\end{figure}

\begin{figure}[hp]
\vspace{-0.1in}
\centerline{
\subfigure[GQRS Performance]{
	\includegraphics[scale=0.3]{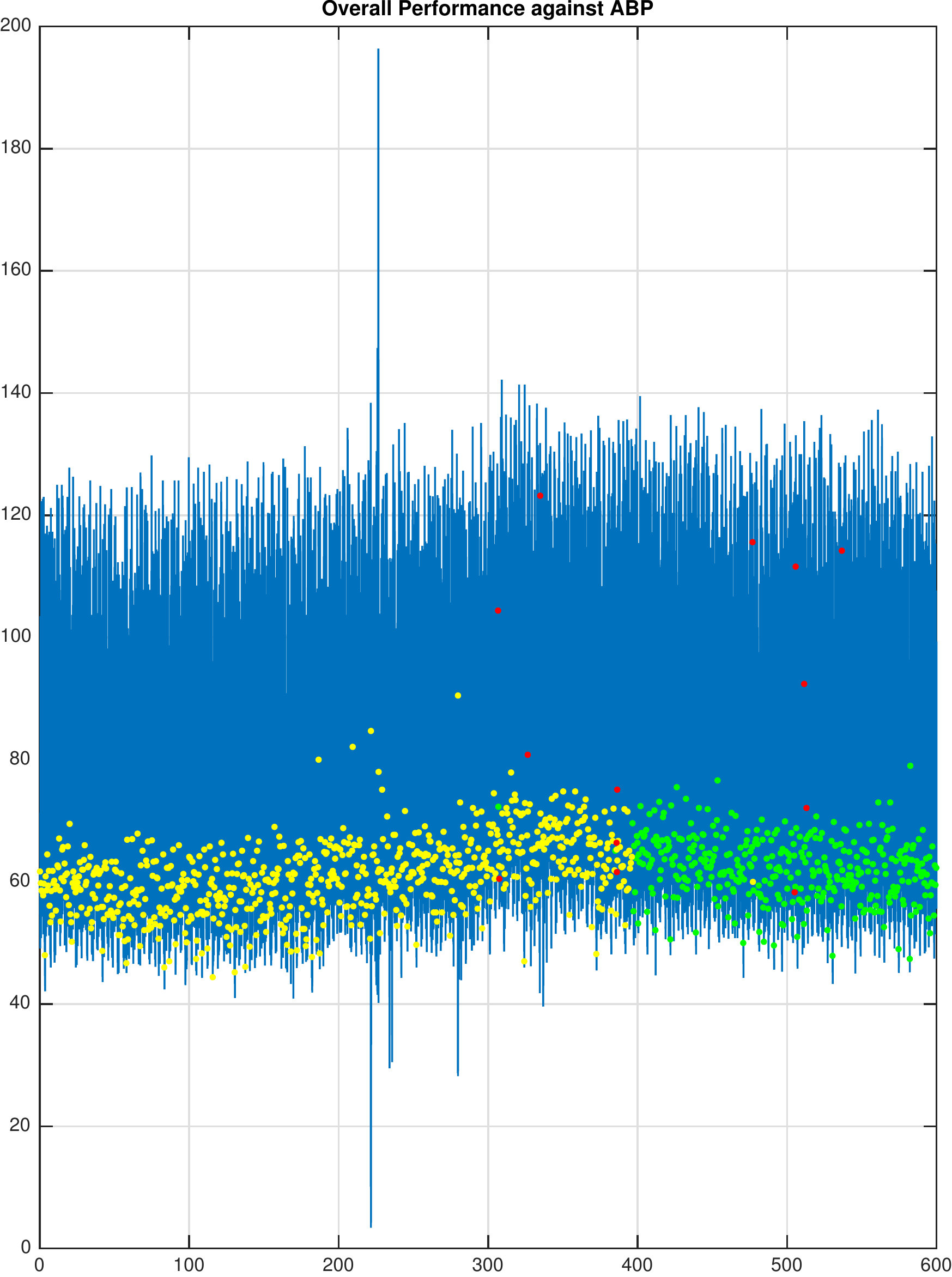}
    \label{GQRS_Performance}
}
\subfigure[DBN-PF Performance]{
    \includegraphics[scale=0.3]{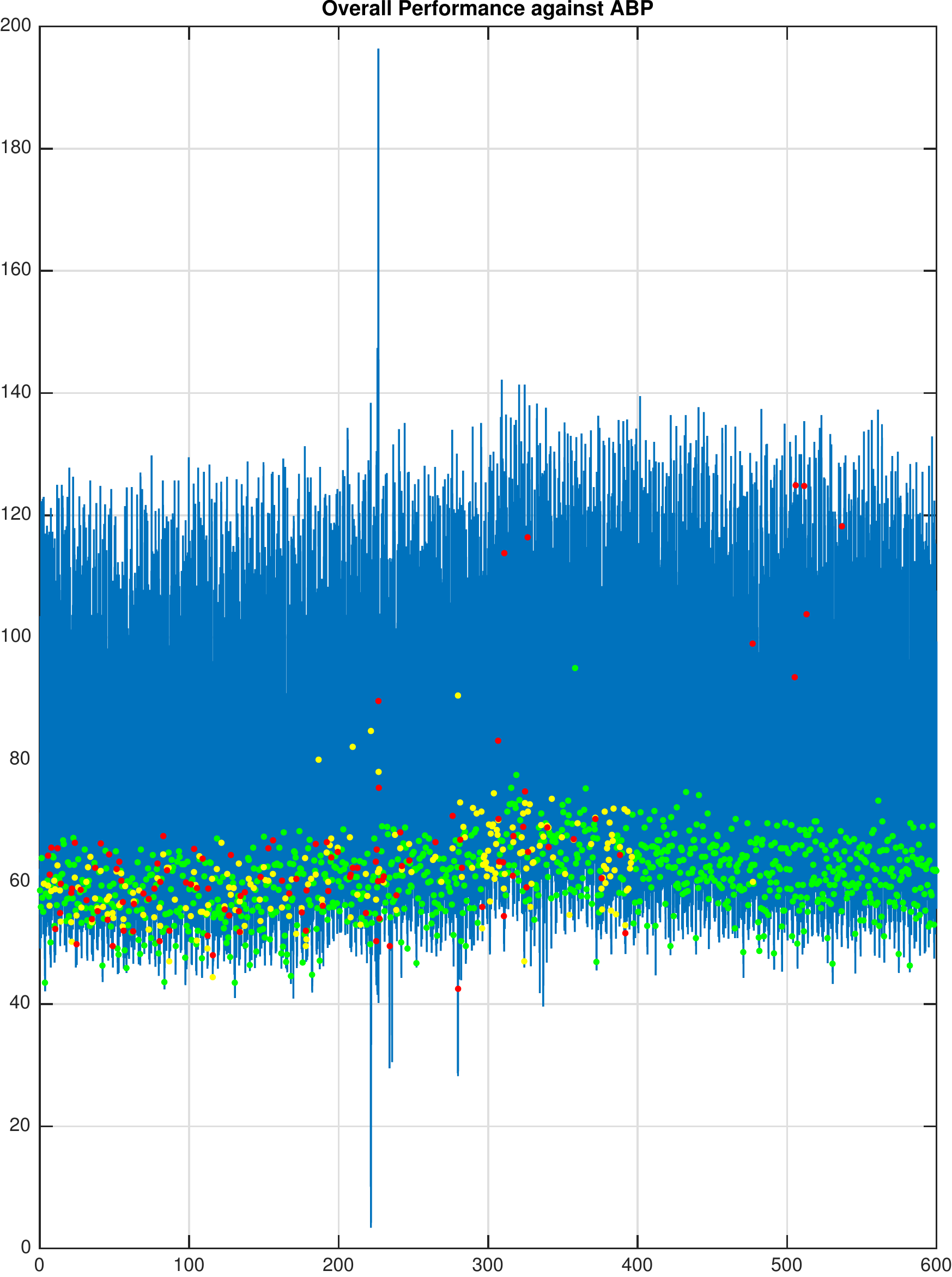}
    \label{PF_Performance}
}
}
\vspace{-0.1in}
\caption{Example 2664 from set-p2.  This visualizes an estimate of the scoring method from Physionet we used.  Green indicates true positives, red is a false positive, and yellow is missed actual annotations.  Note that this visualization is not the same as the method we used to collect the scoring.}
\label{2664_Performance}
\end{figure}

In Figure \ref{2664_quality}, we see that our estimate of the $ECGArtifact$ correlates strongly with the $ECGSQI$.  Wherever the $ECGSQI$ is low, our particle filter correspondingly believes there to be an artifact. The fact that the $ECGArtifact$ variable is quite accurate serves as a proof of concept that our particle filter can track other biometric or sensor state in addition to heart beats.  In addition, we see that the $TrueHR$ matches the actual heart rate quite closely.  This means we can potentially incorporate other variables, such as those that could determine the presence of arrhythmia or other information relevant to the general status of a patient, to create a more refined model.

Then, in Figure \ref{2664_Performance}, we see a strong improvement over GQRS.  Our algorithm has a sensitivity of 0.941 and a predictivity of 0.988, whereas GQRS has a sensitivity of 0.343 and a predictivity of 0.975.  These numbers indicate that for example 2664, GQRS missed a lot of actual beats, but didn't make many false predictions, which is reflected in the figure as well.  Overall, this example highlights the powerful recovery our algorithm elicited simply by combining GQRS and WABP annotations under a physiological model.

\begin{figure}[ht]
	\centerline{
		\includegraphics[scale=0.4]{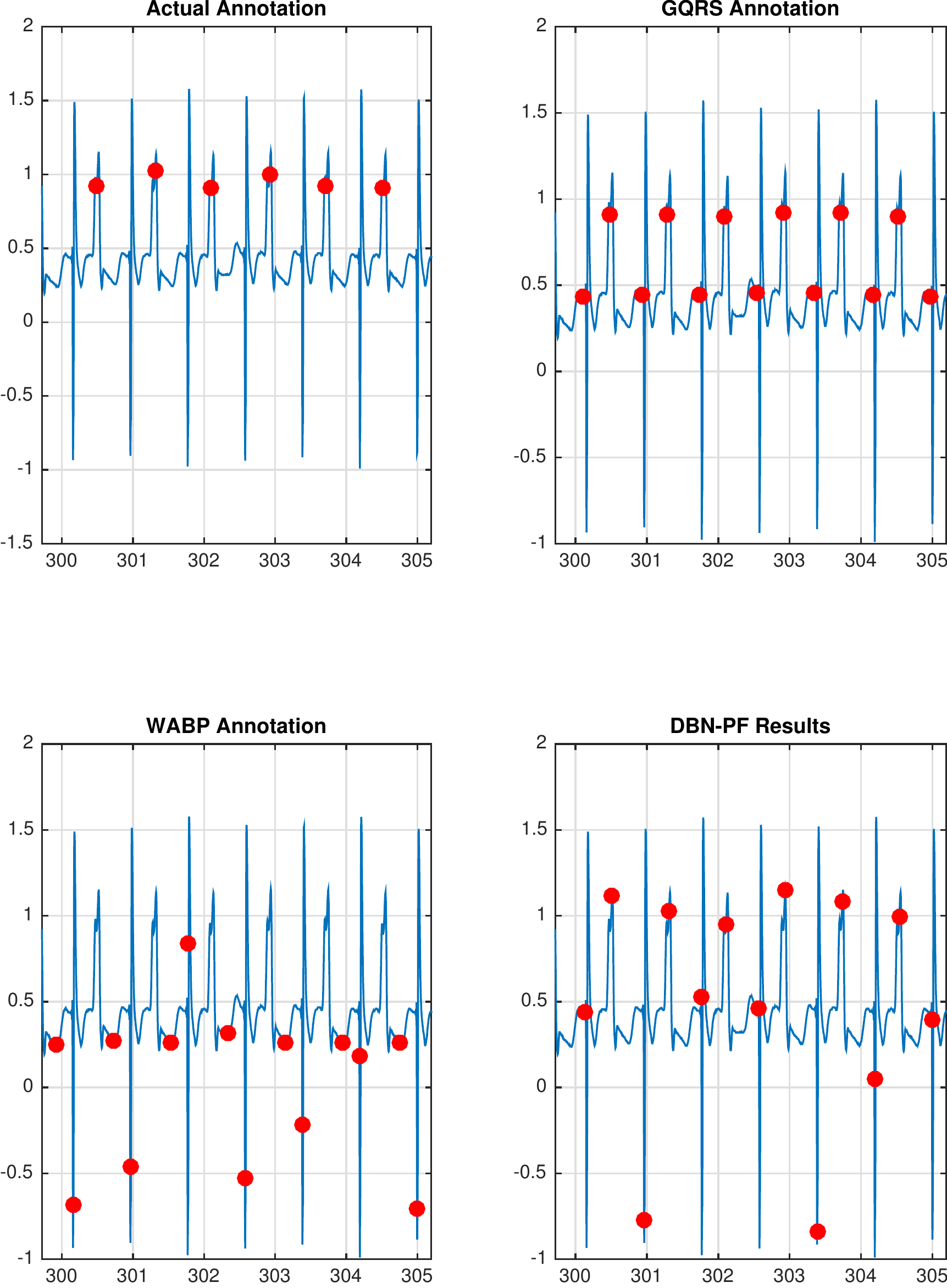}
	}
	\caption{Example 1033 from the set-p2.  The red dots incidate annotations.}
    \label{double_annotation}
\end{figure}

In our last record of interest, 1033 from set-p2, we observe a phenomenon we denoted as "double annotations" pictured in Figure \ref{double_annotation}.  These double annotations are mainly due to artificial pacemakers and low dicrotic notches.  This is where the signals are shaped in such a way that either the GQRS or WABP or both algorithms annotate an extra set of beats.  Here we observe the case where both GQRS and WABP believe there to be a double annotation, resulting in our particle filter believing there to be a double annotations as well.  These double annotation phenomena are the primary reason for our lower positive predictivity in the next results section.  Recovering from double annotations is actually quite difficult within the generative capabilities of our probabilistic model, however through the introduction of new features the double annotations can potentially be ameliorated.  Without other data about the signal, determining the presence of double annotations is infeasible.

\subsection{Results - Overall}
In order to generate these results, we took the top submissions on Physionet, downloaded their entries, and re-ran them against the exact same datasets we used.  We then used Physionet's provided bxb function to compare the generated heart beat annotations with the reference ones and to determine the sensitivity and predictivity for a given record. Finally, we averaged the sensitivity and predictivity across all records from a given dataset.  This was for the sake of consistency in our comparisons.
While all the submissions ran without errors on set-p1 and set-p2, our algorithm, Pangerc's, and Johnson's did not run properly on some of the records in the MGH/MF Waveform Database. We omitted those records when calculating scores.  The results for set-p1 (top left), set-p2 (top right), and the MGH/MF Waveform Database (bottom):
\begin{center}
	\begin{tabular}{||c c c||} 
	    \hline 
	     & set-p1 &  \\
		\hline
		Method & Sensitivity & Pos. Pred. \\ [0.5ex] 
		\hline\hline
		DBN-PF & 0.99893 & 0.99380 \\ 
		\hline\hline
		GQRS & 0.99942 & 0.99320 \\
		\hline
		WABP & 0.39006 & 0.38997 \\
		\hline
		Pangerc & \textbf{0.99995}  & 0.99950 \\
		\hline
		Johnson & 0.99829  & 0.99805 \\
		\hline
		Antink & 0.99965 & 0.99972  \\
		\hline
		De Cooman & 0.99962 & 0.99985  \\ 
		\hline
		Johannesen & 0.99956 & 0.99809 \\
		\hline
		Vollmer & 0.99963 & \textbf{0.99992} \\
		\hline		
	\end{tabular}
	\begin{tabular}{||c c c||} 
	    \hline 
	     & set-p2 &  \\
		\hline
		Method & Sensitivity & Pos. Pred. \\ [0.5ex] 
		\hline\hline
		DBN-PF & 0.92808 & 0.89527 \\ 
		\hline\hline
		GQRS & 0.88134 & 0.85048 \\
		\hline
		WABP & 0.48544 & 0.45173 \\
		\hline
		Pangerc & \textbf{0.95737}  & \textbf{0.94473} \\
		\hline
		Johnson & 0.92734  & 0.89265 \\
		\hline
		Antink & 0.91394 & 0.91829  \\
		\hline
		De Cooman & 0.88364 & 0.88341  \\ 
		\hline
		Johannesen & 0.91535 & 0.86260 \\
		\hline
		Vollmer & 0.91092 & 0.91334 \\
		\hline		
	\end{tabular}
	\begin{tabular}{||c c c||} 
	    \hline 
	    MGH/MF & Waveform & Database \\
		\hline
		Method & Sensitivity & Pos. Pred. \\ [0.5ex] 
		\hline\hline
        DBN-PF & 0.92853 & \textbf{0.93716} \\ 
		\hline\hline
		GQRS & 0.80400 & 0.85934 \\
		\hline
		Pangerc & 0.88567  & 0.91749 \\
		\hline
		Johnson & \textbf{0.93832}  & 0.91808 \\
		\hline
	\end{tabular}
\end{center}

Overall, the performance from GQRS is nigh impossible to beat on set-p1, precisely because the data is so clean and regular.  It seems highly unlikely that there is any statistical significance to be derived from set-p1.  All algorithms perform extremely well, except for WABP which suffers primarily from delay.

In regards to set-p2, we start to observe some differentiation.  Comparing against the other entries, we see that our particle filter ends up outperforming all algorithms in regards to sensitivity except for Pangerc on set-p2.  Our predictivity does suffer due to double annotations, but we still perform well overall.  This is fairly strong evidence that our algorithm is capable of accurately combining the information from multiple channels of signals. In set-p2 we see that the performance of the Pangerc submission is substantially better than the others.  This is likely due to their use of custom ECG and BP pulse detectors.  Their QRS detector (repdet) provided a much improved performance over GQRS on set-p2 in particular, likely due to their inclusion of a step in the detector to identify double annotations (paced beats) \cite{PMEA} \cite{PangercPaper}.  This step may account for Pangerc's improved performance over the algorithms that used GQRS.

Finally, for the MGH/MF Waveform Database, we note that the particle filter outperforms all other algorithms in predictivity, and does very well in the sensitivity aspect as well with approximately .929 sensitivity and .937 predictivity.  In comparison to Pangerc and GQRS this is a great improvement, and in comparison to Johnson, we improve predictivity and only slightly lose out on sensitivity.  Our algorithm was able to perform well on multiple datasets, which suggests that the improvement was fairly significant.

\section{Discussion}
In this section, we discuss some of the interesting features of our algorithm.  First, it is important to state that our algorithm is slower than other standard signal detection algorithms mainly because it is a simulation based filtering algorithm.  In general, on the set-p1 (10 minutes signals), our algorithm takes 94.33 seconds on average to run on MATLAB r2015a (on a MacBook Pro with a 2.9 GHz Intel Core i5 processor and 8 GB 1867 MHz DDR3 memory), but it is worth noting that it is quite feasible for it to be implemented as an online algorithm.  This is simply because particle filters process data sequentially.

We showed an example of double annotation in the previous section.  In order to mitigate double annotations we would have to augment our model.  One way to do this would be to incorporate information about the amplitude of the beat as an observation to determine if our algorithm should expect a double annotation, and determine peaks based off the new information.  The model augmentations would likely be a few variables to probabilistically indicate the presence of double annotations and represent the amplitude of the signal.

Because there were so many other meaningful algorithms, it is worth comparing our own algorithm against the other top submissions.  First of all, our algorithm inherently differed from the other approaches, and was the only one that focused on performing probabilistic inference on a Dynamic Bayesian Model.  In terms of similarities, our algorithm slightly resembles the Johnson algorithm, because both algorithms use signal quality indices.  However, not only do they focus on a deterministic switching, in the latter part of their algorithm they focus on the regularity of signals besides ECG and ABP as well.  As a whole, our algorithm outperforms the others that relied on implementing a form of intelligent switching.  Only Pangerc, who relied on their reimplemented beat detection algorithm (repdet) outperformed our own \cite{PMEA}.

Another point of discussion is actually a point of differentiation between our algorithm and the others, which is our flexibility.  With the flexibility of inputs, it becomes feasible to consider using alternate detectors to further strengthen our algorithm.  One alternative would be to use the detectors used in the Pangerc submission, which should in theory greatly boost our performance.

Apart from performance concerns, we can consider the ease of implementation.  Model based probabilistic inference approaches are becoming more and more appealing due to the  emergence of probabilistic programming languages (PPL). A probabilistic programming language is a high-level language that makes it easy to represent probabilistic models and perform inference over them. PPLs enable domain experts who don't have enough experience in probability theory or machine learning to use state-of-the-art machine learning methodologies to perform meaningful inferences \cite{gordon2014probabilistic}. Because the problem we are analyzing is within the domain of physicians and clinicians, it would be best if we could empower them to create the physiological models they wished to represent on their own.  As a proof of concept, we have also implemented our probabilistic model in a probabilistic modeling language called BLOG \cite{BLOGPaper}.  We used BLOG's particle filtering engine to perform the state estimation, and our results agree with our MATLAB implementation.  For physicians, they simply need to implement their model, and use the particle filter (or any other estimation technique) implemented in BLOG. For a glimpse at the BLOG implementation, refer to Appendix \ref{BLOG Code}.

\section{Conclusion}
Based on our results, we find that using particle filtering on our DBN model performs quite well.  It matches or outperforms a majority of the top submissions for the Physionet 2014 Challenge.  In regards to the default Physionet beat detector GQRS, we improve in set-p2 by about $4\%$ in both sensitivity and positive predictivity.  As a whole, our particle filter serves as a strong proof of concept that a probabilistic model based inference approach can robustly detect heart beats.

Our probabilistic model is not completely perfect, but some of our algorithm's advantages include: 
\begin{enumerate*}
	\item Utilizes multi-channel information by incorporating the ECG-ABP peak delay (latency).
	\item Flexible model - easy to incorporate new variables and relationships.
	\item Flexible observations - easy to switch out GQRS and WABP for any other signal processing algorithms.
	\item Consistent with a physiological perspective.
	\item Can be implemented using a PPL.
	\item Unlike many machine learning techniques, our model does not require a training set.
\end{enumerate*}  Some of our disadvantages include:
 \begin{enumerate*}
	\item Slower run time.  One ten-minute signal takes approximately a minute and a half for the algorithm to run.
	\item Depends on the signal processing methods heavily.
\end{enumerate*}

In terms of future work on the particle filter, we can incorporate the preprocessing mentioned in the other submissions for the sake of reducing noise.  In addition, our algorithm only looks at one ECG and one ABP signal, when there are a multitude of other signals that could be utilized.  Incorporating these would likely improve performance, as would initializing the delay based on one of the methods described in the other submissions.  In the far future, this algorithm could be tweaked to directly detect arrhythmia.  By placing emphasis on the artifacts and introducing arrhythmia variables, it would be possible to develop beliefs regarding these cardiac abnormalities.  Since we have improved the detection rate for heart beats, it stands to reason we can improve the detection rate for the arrhythmia as well as for other problems that can be modelled physiologically.

\section{Acknowledgements}
We are thankful to Xiao Hu, Quan Ding, Yong Bai, Rebeca Salas-Boni and Daniel Schindler for helpful discussions.

\nocite{*}
\bibliographystyle{dcu}
\bibliography{paper}

\newpage
\clearpage
\appendix

\begin{center}
\LARGE{Supplemental Material}
\vspace{3em}
\end{center}
\section{Propagation Model} \label{Propagation Model Supplemental}
The propagation model (transition model) encodes interdependent evolution of our relevant physiological variables over time.  The structure of the dependencies is observable in the figure in section \ref{DynamicBayesNet}.

\subsection{RestHR}
$RestHR$ is a real-valued static variable that represents the resting heart of a particular patient, which is simply the patient's expected heart rate while at rest.  The prior value is a Gaussian around the average heart rate as determined by patient demographics.  It obeys the following static (i.e., identity) propagation function:
\begin{equation}
	RestHR_{t+1} \leftarrow RestHR_t
\end{equation}
\subsubsection{Latency}
\begin{figure}[ht]
	\centerline{
		\includegraphics[scale=.5]{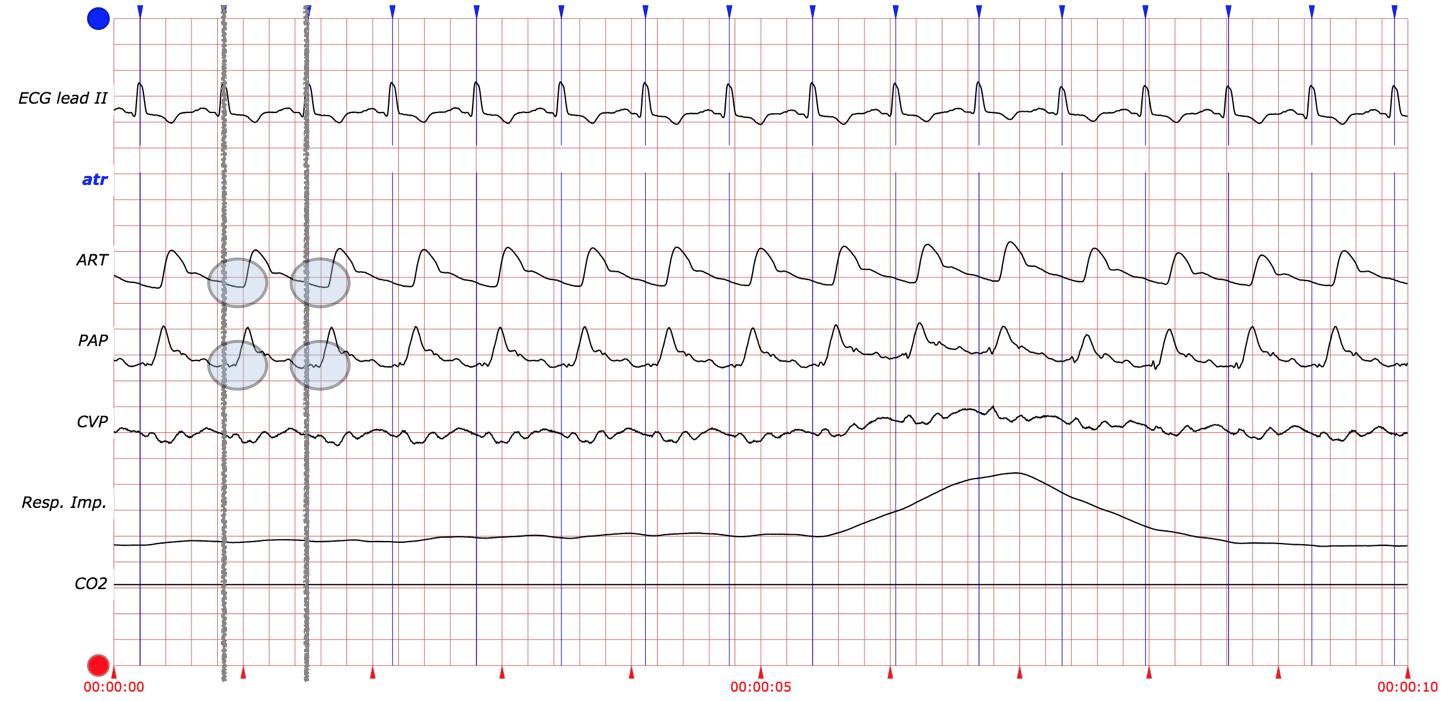}
	}
	\caption{The blue lines represent heart beat annotations.  The gray lines and circles are simply to highlight the annotated heart beat and the delay present in ART and PAP (the ABP signals).}
	\label{LatencyExample}
\end{figure}
In Figure \ref{LatencyExample}, it is possible to see that the heartbeats fall on, or very near to, the R peaks in the ECG's QRS signals.  Conversely, in the ABP signals the heart beat does not fall on the peaks pictured, but instead it falls a reliable distance before the peaks.  In our model, $Latency$ is an integer-valued static variable that represents this relationship between heart beats and the blood pressure.  The prior value is a Gaussian around 200 milliseconds.  It obeys the following propagation function:
\begin{equation} 
	Latency_{t+1} \leftarrow Latency_t
\end{equation}
\subsection{TrueHR}
$TrueHR$ is a real-valued variable that represents the heart rate of a patient at the current time.  The prior value of the $TrueHR$ starts off as a gaussian around the $RestHR$.  Here $norm$ means sampling from a gaussian distribution with a given $\mu$ and $\sigma$.
\begin{equation} 
	TrueHR_0 \leftarrow RestingHR_0 + 5*norm(\mu = 0,\sigma = 1)
\end{equation}
Since the true heart rate of a patient might vary over a few minutes where the resting heart rate might vary over a few months, the true heart rate is not a static variable.  It obeys the following propagation function:
\begin{equation} 
	TrueHR_{t+1} \leftarrow .8*TrueHR_t + .2*RestHR_t + 15*norm(\mu = 0,\sigma = 1)
\end{equation}
\subsection{ECGPeak}
$ECGPeak$ is a boolean-valued variable that is $1$ if there should be a beat annotated at the current timestep and $0$ otherwise.  $ECGPeak$'s prior starts as $1$ with a small probability (currently $.01$).  It obeys the following propagation function:
\begin{equation} 
	ECGPeak_{t+1} \leftarrow Bernoulli(P)
\end{equation}
We calculate $P$ dynamically.  As time progresses, the $P$ parameter will change depending on the current values of the $TrueHR$ and $ECGLastPeak$.  First we define the difference between the current timestep and the lastpeak as $diff = t-ECGLastPeak_t$.  Then we define $BeatWindow_t=60/(window*TrueHR_t)$, which is the number of windows per beat based on the current heart rate.
\begin{equation}
	\begin{split}
		P = binopdf(& x=max(mod(diff,BeatWindow_t), mod(diff,BeatWindow_t) + BeatWindow_t), \\ 
		& n=3/2*BeatWindow_t, p=2/3)
	\end{split}
\end{equation}
\begin{figure}
	\centerline{
		\includegraphics[scale=.5]{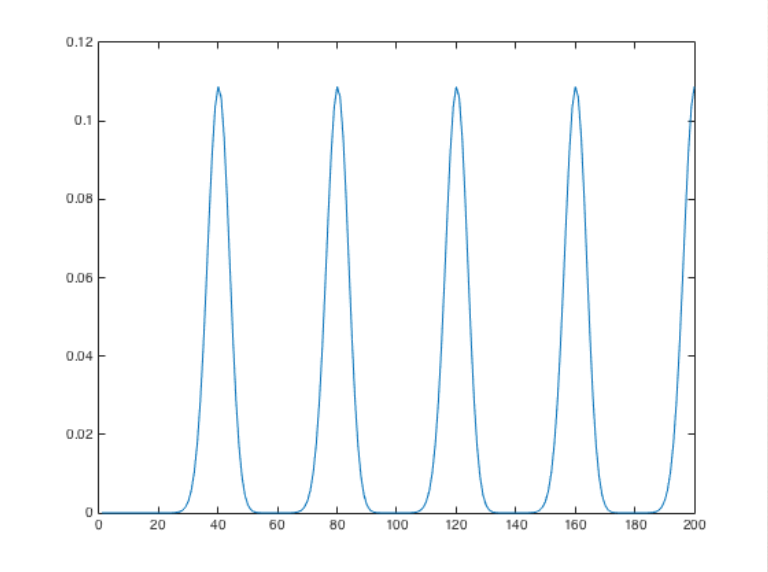}
	}
	\caption{This graph represents the value of $P$ as a function of $diff$ for a fixed TrueHR and window.  This is an example with a TrueHR of 60 bpm and a window of .025 s.}
	\label{MultiBino}
\end{figure}
Thus, we represent the probability according to repeated binomial distributions, as in figure \ref{MultiBino}.  The reason for this calculation is to create a few important properties.  The first is to create a memory that allows us to believe beats should occur based on the $TrueHR$ and to not preclude beats even if we don't believe there to be a beat earlier on.  It also has the convenient property of generally not allowing us to double annotate beats, because it's unlikely the heart will beat twice in a short span of time.  The other reason is that the binomial distributed random variable $X$ with parameters $n$ and $p$ has an $E[X]=np$ and $Var[X]=np(1-p)$.  This means that we can control the expected value in our case to be $BeatWindow$ and the variance to be $1/3*BeatWindow$, using our values of $n$ and $p$.
\subsection{ECGLastPeak}
$ECGLastPeak$ is an integer-valued variable that represents the last time we believed there was a peak based on ECG.  The prior for $ECGLastPeak$ is a uniform distribution in the range of integers between [$-BeatWindow,-1$].  The $ECGLastPeak$ helps to shift the expected location of all of the heartbeats, because of the multimodal value of $P$ in the $ECGPeak$.  The uniform sampling means that the prior believes the expected heartbeats may be shifted to cover all possible initializations.  It obeys the following propagation function:
\begin{displaymath}
	ECGLastPeak_{t+1} = \left\{
	\begin{array}{lr}
		t + 1 & if ECGPeak_{t+1} == 1\\
		ECGLastPeak_{t} & if ECGPeak_{t+1} == 0
	\end{array}
	\right.
\end{displaymath} 
In words, this equation states that the $ECGLastPeak$ will change purely to record the last time $ECGPeak$ was 1.
\subsection{ABPPeak}
$ABPPeak$ is a boolean-valued variable that represents whether there is a peak based on the $ABP$ and the $ECG$.  Since theoretically the $ABPPeak$ should align with the $ECGPeak$ while accounting for the $Latency$, it obeys the following propagation function (even for the prior):
\begin{equation} 
	ABPPeak_{t+1} \leftarrow (t == ECGLastPeak_t + ParticleMean(Latency_t))
\end{equation}
Here, $ParticleMean$ means that we take the mean across all the particles.  This is admittedly unorthodox, but it means that we ameliorate the issue of double and triple annotations in adjacent locations for the $ABPPeak$.  If the particle filter had a while to learn the latency and let it converge, then this fix would be unnecessary, but, as is, this fix means that the particle filter's particles will not overly diverge at the $ABPPeak$.
\subsection{ABPLastPeak}
$ABPLastPeak$ is an integer-valued variable that represents the last time we believed there was a peak based on both ECG and ABP.  The prior starts off as $ECGLastPeak_0+Latency_0$.  It obeys the following propagation function:
\begin{displaymath}
	ABPLastPeak_{t+1} = \left\{
	\begin{array}{lr}
		t + 1 & if ABPPeak_{t+1} == 1\\
		ABPLastPeak_{t} & if ABPPeak_{t+1} == 0
	\end{array}
	\right.
\end{displaymath} 
\subsection{ECGArtifact and ABPArtifact}
$ECGArtifact$ and $ABPArtifact$ are boolean-valued variables that represents whether we believe there is currently an artifact associated with either $ECGPeak$ or $ABPPeak$, respectively.  Their prior and propagations behavior is exactly the same.  $1$ represents an artifact and $0$ represents no artifact.  The prior belief for the artifacts starts off as $1$ with a small probability (currently $.01$) It obeys the following propagation function:
\begin{equation} 
	Artifact_{t+1} \leftarrow Bernoulli(P_A)
\end{equation}
Here, $P_A$ represents the Pr($Artifact_{t+1}|Artifact_t$), which is determined by the following conditional probability table:
\begin{center}
	\begin{tabular}{||c c c||} 
		\hline
		$Artifact_t$ & $Artifact_{t+1}$ & Pr($Artifact_{t+1}|Artifact_t$) \\ [0.5ex] 
		\hline\hline
		1 & 1 & 0.99 \\ 
		\hline
		1 & 0 & 0.01 \\
		\hline
		0 & 1 & 0.01 \\
		\hline
		0 & 0 & 0.99 \\ [1ex] 
		\hline
	\end{tabular}
\end{center}
This table represents a form of inertia.  Given that there is currently an artifact, the probability that there continues to be an artifact is high.  Likewise, it there is currently an absence of an artifact, the probability that there continues to be an absence is similarly high.
\section{Observation Model} \label{Observation Model Supplemental}
The observation model (sensor model) encodes our beliefs about the functions we use to derive observations and the probability that the observations correspond to the current states.  The structure of the dependencies is observable in the figure in section \ref{DynamicBayesNet}.

\subsection{Annotation Observations}
These are the $ABPAnn$ and the $ECGAnn$ variables in the model.  The observations are derived by using the WABP and GQRS algorithms that are provided by Physionet.  The algorithms simply use a specified window size and see if the signal processing algorithms place any annotations within a given window.  If so, then the $Ann$ observation is set to be true ($1$), and otherwise it's set to be false ($0$).  Then, we define the probability of the annotations given the states according to the following probability table:
\begin{center}
	\begin{tabular}{||c c c c||} 
		\hline
		$Peak$ & $Artifact$ & $Ann$ & Pr($Ann|States$) \\ [0.5ex] 
		\hline\hline
		0 & 1 & 1 & $NormBeatProb$ \\ 
		\hline
		0 & 1 & 0 & $1-NormBeatProb$ \\
		\hline
		0 & 0 & 1 & $BeatProb$ \\
		\hline
		0 & 0 & 0 & $1-BeatProb$ \\
		\hline
		1 & 1 & 1 & 0.7 \\ 
		\hline
		1 & 1 & 0 & 0.3 \\
		\hline
		1 & 0 & 1 & 0.99 \\
		\hline
		1 & 0 & 0 & 0.01\\
		\hline
	\end{tabular}
\end{center}
In the case that there is a peak and there is no artifact, then the belief that there should be an annotation is quite high ($.99$), because there is high likelihood of signal accuracy.  Correspondingly the belief that there should not be an annotation is quite low ($.01$).

In the case that there is a peak and there is an artifact, then the belief that there should be an annotation is lower than without an artifact ($.7$), because the artifact means that we are not entirely able to trust the signal.  Correspondingly the belief that there should not be an annotation is higher than without an artifact ($.3$).  In general when we fix the other states, the presence of an artifact makes our annotation beliefs less certain.  

Next in the case that there is no peak and no artifact, $BeatProb$ should represent the likelihood that the observation can be trusted.  Since the probability of the annotation occuring in this case truly depends on the $LastPeak$ and the $TrueHR$, we end up calculating it in the same way as the $P$ for the $ECGPeak$.  First we define the difference between the current timestep and the lastpeak as $diff = t-LastPeak$.  Then we define $BeatWindow=60/(window*TrueHR)$, which is the number of windows per beat based on the current heart rate.
\begin{equation}
	\begin{split}
		BeatProb = binopdf(& x=max(mod(diff,BeatWindow), mod(diff,BeatWindow) + BeatWindow), \\ 
		& n=3/2*BeatWindow, p=2/3)
	\end{split}
\end{equation}
Then, we know that the probability of the annotation not occuring will simply be $1-BeatProb$.

Finally in the case that there is no peak and an artifact, we can apply what we used earlier and say that the presence of an artifact makes our belief about the annotation less certain.  So if we strongly believe there would be an annotation, in the case of an artifact, we only moderately believe there would be an annotation.  Likewise if we don't believe there would be an annotation, then an artifact would make us moderately not believe in an annotation.  This means that we can simply represent the $NormBeatProb$ according to the following definition:
\begin{equation} 
	NormBeatProb = mean(.5, BeatProb)
\end{equation}

\section{Heart Rate Observations}
These are the $WABPHR$ and the $GQRSHR$ variables in the model (collectively called $HRobs$ variables).  The observations are derived by using the WABP and GQRS algorithms that are provided by Physionet.  The algorithms simply use a specified window size and have a sliding window that computes the local heart rate.  The $HRobs$ observations are set accordingly.  Then, we define the probability of the observations given the states according to the following:
\begin{equation} 
	Pr(HRobs|States) = normpdf(TrueHR,HRobs,1/4*HRobs)
\end{equation}
Here, $normpdf$ corresponds to the probability density function of a Gaussian random variable with the given parameters. 

\section{SQI Observations}
These correspond to the $ECQSQI$ and $ABPSQI$ variables.  $ECGSQI$ is calculated by taking two signal processing algorithms and comparing them beat by beat.  It was derived using the ecgsqi function that was developed in another Physionet submission \cite{JohnsonPaper}.  In particular, we use gqrs (unpublished algorithm optimized for sensitivity) and wqrs (open source algorithm optimized for adult human ECGs).  The number of beats that match between the two algorithms is reported as a real number between $0$ and $1$.  $ABPSQI$ is calculated by checking to see if the pressure, the mean arterial pressure, the heart rate, the pulse pressure, and a variety of other physiologic details are within normal ranges or not.  If everything is in a normal range, $ABPSQI$ is $1$, otherwise it is $0$.  This was also derived from the abpsqi function in the same Physionet submission, which was in turn borrowing from other publications \cite{SunPaper} \cite{Sunmastersthesis}. 

The $SQI$ observations are used to choose which observations to depend on.  If the $ECGSQI<.8$ and $ABPSQI==1$, then we weight based on the ABP annotations and heart rates, otherwise we weight based on the ECG annotations and heart rates.

\subsection{BLOG Code - Propagation Functions} \label{BLOG Code}

The following code is the propagation model represented in the BLOG language:

\begin{verbatim}
// Functions
random Real Rest_HR(Timestep t) ~
    if t == @0 then
        Gaussian(avg_hr, 10)
    else
        Rest_HR(prev(t));

random Real True_HR(Timestep t) ~
    if t == @0 then
        Gaussian(Rest_HR(t), 5)
    else
        Gaussian(.2*Rest_HR(prev(t)) + .8*True_HR(prev(t)), 1);

random Integer ECG_Art(Timestep t) ~
    if t == @0 then 
        Bernoulli(.01)
    else case ECG_Art(prev(t)) in {
        0 -> Bernoulli(.01), 
        1 -> Bernoulli(.99)
    };

random Integer ECG_Peak(Timestep t) ~
    if t == @0 then
        Bernoulli(.01)
    else
        Bernoulli(binompdf(round(60/(w_off*True_HR(prev(t)))), 2.0/3.0,
            (toInt(t)-ECG_Last_Peak(prev(t)))%round(60/(w_off*True_HR(prev(t))))));

random Integer ECG_Last_Peak(Timestep t) ~
    if t == @0 then
        UniformInt(-round(60/(w_off*True_HR(t))), -1)
    else
        if ECG_Peak(t) == 1 then
            toInt(t)
    else
        ECG_Last_Peak(prev(t));

random Integer ABP_Art(Timestep t) ~
    if t == @0 then 
        Bernoulli(.01)
    else case ABP_Art(prev(t)) in {
        0 -> Bernoulli(.01), 
        1 -> Bernoulli(.99)
    };

random Integer ABP_Peak(Timestep t) ~
    if (ECG_Last_Peak(t) + Latency) == toInt(t) then
        1
    else
        0;

random Integer ABP_Last_Peak(Timestep t) ~
    if t == @0 then
        ECG_Last_Peak(t) + Latency
    else
        if ABP_Peak(t) == 1 then
            toInt(t)
    else
        ABP_Last_Peak(prev(t));
\end{verbatim}

\subsection{BLOG Code - Observation Functions}

The following code is the observation model represented in the BLOG language:

\begin{verbatim}
// Functions
random Integer ECG_Ann(Timestep t) ~
    if ECG_Peak(t) == 1 then
        if ECG_Art(t) == 1 then
            Bernoulli(.7)
        else
            Bernoulli(.9)
    else
        if ECG_Art(t) == 1 then
            Bernoulli((binompdf(round(60/(w_off*True_HR(t))), 2.0/3.0, 
                (toInt(t)-ECG_Last_Peak(t))%round(60/(w_off*True_HR(t)))) + .5)/2)
        else
            Bernoulli(binompdf(round(60/(w_off*True_HR(t))), 2.0/3.0, 
                (toInt(t)-ECG_Last_Peak(t))%round(60/(w_off*True_HR(t)))));

random Real ECG_HR(Timestep t) ~
    Gaussian(True_HR(t), abs(True_HR(t))/4);

random Integer ABP_Ann(Timestep t) ~
    if ABP_Peak(t) == 1 then
        if ABP_Art(t) == 1 then
            Bernoulli(.7)
        else
            Bernoulli(.9)
    else
        if ABP_Art(t) == 1 then
            Bernoulli((binompdf(round(60/(w_off*True_HR(t))), 2.0/3.0, 
                (toInt(t)-ABP_Last_Peak(t))%round(60/(w_off*True_HR(t)))) + .5)/2)
        else
            Bernoulli(binompdf(round(60/(w_off*True_HR(t))), 2.0/3.0, 
                (toInt(t)-ABP_Last_Peak(t))%round(60/(w_off*True_HR(t)))));

random Real ABP_HR(Timestep t) ~
    Gaussian(True_HR(t), abs(True_HR(t))/4);
\end{verbatim}

\subsection{Running BLOG}
The following is the shell script we ran to execute the blog particle filter.
\begin{verbatim}
#!/bin/sh

time ./../../blog/dblog \
  prop_fn.dblog \
  query.dblog \
  obs_fn.dblog \
  obs.dblog \
  -n 1000 -o out.json
\end{verbatim}

The query.dblog and obs.dblog files contained the data we wanted to load.  This command ran the particle filter with 1000 particles and put the output into the out.json file.  Note: we will be making the Matlab and BLOG code we wrote available at a later date.

\end{document}